\documentclass[10pt,twocolumn,letterpaper]{article}
\usepackage[accsupp]{axessibility}

 \usepackage{cvpr}              
\usepackage[dvipsnames]{xcolor}

\definecolor{cvprblue}{rgb}{0.21,0.49,0.74}
\usepackage[pagebackref,breaklinks,colorlinks,citecolor=cvprblue]{hyperref}
\usepackage{float}
\usepackage{url}
\def\paperID{*****} 
\def\confName{CVPR}
\def\confYear{2024}

\title{DeepLocalization: Using change point detection for Temporal Action Localization}

\author{Mohammed Shaiqur Rahman\\
Iowa State University\\
shaiqur@iastate.edu
\and
Ibne Farabi Shihab\\
Iowa State University\\
ishihab@iastate.edu
\and 
Lynna Chu\\
Iowa State University\\
lchu@iastate.edu
\and
Anuj Sharma\\
Iowa State University\\
anujs@iastate.edu
}

\begin{document}
\maketitle
\begin{abstract}
In this study, we introduce DeepLocalization, an innovative framework devised for the real-time localization of actions tailored explicitly for monitoring driver behavior. Utilizing the power of advanced deep learning methodologies, our objective is to tackle the critical issue of distracted driving—a significant factor contributing to road accidents. Our strategy employs a dual approach: leveraging Graph-Based Change-Point Detection for pinpointing actions in time alongside a Video Large Language Model (Video-LLM) for precisely categorizing activities. Through careful prompt engineering, we customize the Video-LLM to adeptly handle driving activities' nuances, ensuring its classification efficacy even with sparse data. Engineered to be lightweight, our framework is optimized for consumer-grade GPUs, making it vastly applicable in practical scenarios. We subjected our method to rigorous testing on the SynDD2 dataset, a complex benchmark for distracted driving behaviors, where it demonstrated commendable performance—achieving 57.5\% accuracy in event classification and 51\% in event detection. These outcomes underscore the substantial promise of DeepLocalization in accurately identifying diverse driver behaviors and their temporal occurrences, all within the bounds of limited computational resources.

\end{abstract}    
\section{Introduction}
\label{sec:intro}

According to the National Highway Traffic Safety Administration (NHTSA), distracted driving was responsible for 3,522 fatalities in 2021, alongside causing injuries to an additional 362,415 individuals in motor vehicle accidents \cite{noauthor_distracted_2023}. Distracted driving encompasses any activity that diverts attention from the primary task of driving safely. This includes but is not limited to engaging in phone conversations or texting, eating and drinking, interacting with passengers, and adjusting the vehicle's stereo, entertainment, or navigation systems. Essentially, any action that detracts from the focus on safe driving falls under this category \cite{noauthor_distracted_2023}. Drivers often engage in secondary activities to avoid feeling drowsy; however, these actions lead to distractions.

The consequences of distraction are significant, often resulting in crashes, some of which are fatal. Substant economic repercussions exist beyond the physical harm inflicted on those involved in accidents. The NHTSA has reported that crashes incurred a total cost of 340 billion dollars in 2019. Specifically, distraction-related crashes contributed approximately 98.2 billion dollars to this figure, representing 29\% of the total traffic-related economic costs. As such, there is a pressing need to mitigate the damages caused by distracted driving and enhance road safety. One viable solution is the implementation of a real-time driver behavior monitoring system. Such a system would alert drivers upon detecting any form of distraction. Given their low maintenance and cost-effectiveness, in-vehicle cameras are optimal for real-time driver behavior monitoring \cite{venkatachalapathy_deep_2023}.

In recent years, the surge in computational capabilities, coupled with advancements in deep learning algorithms, has significantly bolstered the prominence of computer vision technologies, especially in the realm of driver behavior classification. Unlike traditional machine learning algorithms—which rely on manually defined key features and specific classifiers for image classification, a process that varies across different datasets and complicates application in diverse scenarios—the advent of deep learning techniques, such as Convolutional Neural Networks (CNNs), has marked a transformative shift. CNNs possess the unique ability to autonomously extract features from any given input data, thereby achieving superior accuracy in both the detection and classification processes. This evolution has spurred extensive research, yielding several CNN models like ResNet50 \cite{he_deep_2016}, AlexNet \cite{krizhevsky2012imagenet}, and InceptionV3 \cite{szegedy_rethinking_2016}, among others, that stand out for their enhanced performance in image classification tasks. Consequently, these models often serve as foundational or baseline models for researchers embarking on image classification or detection projects tailored to their respective projects' specific requirements and constraints.

Parallel to these technological advances, the field has seen an uptick in naturalistic driving studies (NDS), aimed at examining distracted driver behavior within a realistic driving environment. Such studies are invaluable, as the data garnered can shed light on various aspects of the driver's state, including drowsiness, alertness, and distraction, thus providing deep behavioral insights. Notably, the Second Strategic Highway Research Program Naturalistic Driving Study (SHRP 2) \cite{shrptwo}, undertaken by the Virginia Tech Transportation Institute (VTTI), stands out as the largest of its kind, amassing a staggering 2 Petabytes of driving data. Analysis of the SHRP 2 data revealed that the majority of crashes are closely associated with factors attributed to the driver (such as error, impairment, fatigue, and distraction), constituting nearly 90 of all recorded incidents \cite{dingus2016driver}. However, the utility of NDS is not without challenges, primarily due to susceptibility to noise and other data quality issues. To address these concerns, the introduction of SynDD2 \cite{rahman2023synthetic} in the 7th AI City Challenge \cite{naphade_7th_2023} represents a notable effort to mitigate such data quality problems, paving the way for more accurate and reliable analysis of driver behavior.

Determining the driver's behavior involves recognizing the driver's behavior (activity recognition) and finding the duration of such behavior (TAL- Temporal Activity Localization). The CNN-based models \cite{feichtenhofer_slowfast_2019,feichtenhofer_x3d_2020,xu_videoclip_2021} are used for the former, and for the latter various approaches have been explored such as Graph Convolutional Networks \cite{zeng_graph_2019}, 1D temporal segments \cite{chao_rethinking_2018}, sliding window \cite{wang_temporal_2021}. Although there has been various research for TAL, using a change point algorithm has not been explored. Hence, we propose an approach combining a Graph-Based Change-Point Detection \cite{10.1214/18-AOS1691,10.1214/14-AOS1269} for temporal action localization and video LLM model \cite{maaz2023videochatgpt} for classification. 

The contributions of our work are as follows:
\begin{enumerate}
    \item Exploring a novel approach for finding the start and end times of an activity.  Our approach doesn't require any training, the algorithm only needs the key points to find action proposals.
    \item The overall framework is novel from generating action proposals to action classification. It is lightweight which is optimized for consumer-grade GPUs.
\end{enumerate}

The rest of the paper is organized as follows: Section \ref{sec:lit_review} reviews the existing literature. Sections \ref{sec:methodology} and \ref{sec:experiments} detail our methodology and introduce our proposed approach with results. Finally, in Section \ref{sec:results}, we discuss our findings and suggest avenues for future research.

\section{Related Works}
\label{sec:lit_review}

In the past few years, there has been a significant increase in research on driver action recognition. The main objective of this research is to identify and predict risky driving behaviors that lead to accidents caused by distracted driving. Many researchers are developing models and exploring different methods to achieve this goal, significantly focusing on using supervised learning techniques.

\textbf{Temporal Action Localization:} Temporal Action Localization (TAL) is a dynamic field that identifies the precise start and end times of actions within video segments. This process typically involves the initial generation of numerous candidate segments as action proposals, followed by the classification of these proposals into their respective action categories. To enhance the precision of action localization, previous studies have adopted a variety of strategies, such as refining detection boundaries \cite{lin2018bsn,zhao2021bottomup}, creating frame-wise action labels to delineate the temporal limits of actions \cite{3391283d64aa4a21afcf90e3e9ea07cb}, using graphs structures \cite{xu2020gtad,zeng2019graph} and improving the quality of action proposals through the aggregation of context at both the boundary and proposal levels \cite{chen2021dcan}. Despite the broad application of these methods across different research areas, there has been limited exploration into the use of graph-based change point detection algorithms for generating action proposals. The innovative work presented in \cite{10.1214/18-AOS1691} showcases the application of such an algorithm to identify intervals within data where there is a significant shift in distribution, a technique that has shown promise in the context of action proposal identification in videos.

\textbf{Video recognition using LLM:} The integration of visual models with Large Language Models (LLMs) has garnered significant attention following the success of ChatGPT. This emerging field has seen remarkable advancements, notably the pioneering work demonstrated by Yuan et al. \cite{yuan2021florence}, which laid the groundwork for subsequent innovations. Among these, the Querying Transformer proposed by Li et al. \cite{li2023blip2} stands out for its innovative approach to mapping images into the text embedding space, enhancing the interaction between visual data and language models. Further developments by Liu et al. \cite{liu2023visual} and Dai et al. \cite{dai2023instructblip} have introduced methods that facilitate visual conversation, expanding the capabilities of LLMs in understanding and generating responses based on visual inputs. In our research, we have chosen to focus on the work by Maaz et al. \cite{maaz2023videochatgpt} for activity recognition within video data. This approach is distinguished by its adapted LLM framework, incorporating the visual encoder from CLIP \cite{radford2021learning} alongside the Vicuna language decoder \cite{zheng2023judging}. This combination not only leverages the strengths of both models but also addresses the challenge of capturing temporal dynamics and ensuring video-to-frame coherence. Consequently, it offers an optimal solution for analyzing activities in video content, especially in scenarios where the duration of activities varies unpredictably across the dataset.
\begin{figure}[H]
  \centering
   \includegraphics[width=0.8\linewidth]{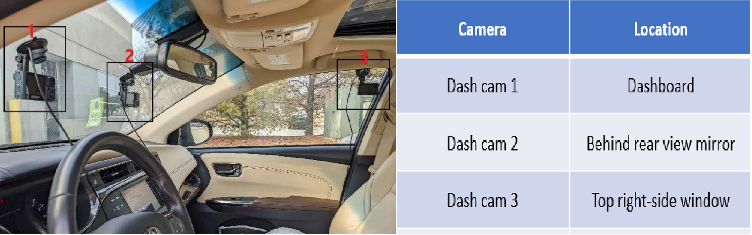}

   \caption{Showing camera positions \cite{RAHMAN2023108793}}
   \label{fig:camerpos}
\end{figure}
\begin{figure}[!htb]
    \centering
    \begin{subfigure}{0.15\textwidth}
        \centering
        \includegraphics[width=\linewidth,height=3cm]{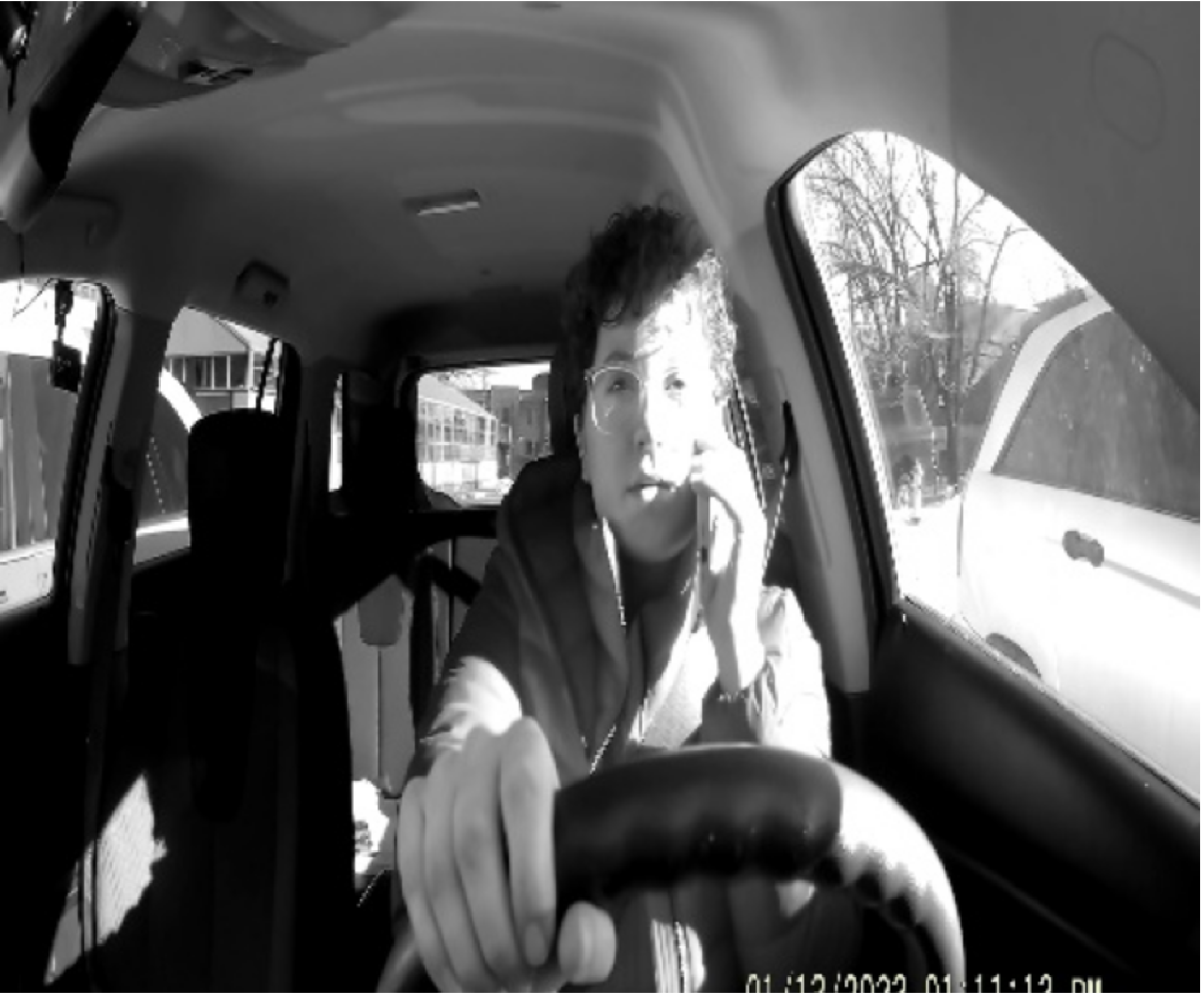}
        \caption{Dashboard}
        \label{fig:sub1}
    \end{subfigure}%
    \hfill
    \begin{subfigure}{0.15\textwidth}
        \centering
        \includegraphics[width=\linewidth,height=3cm]{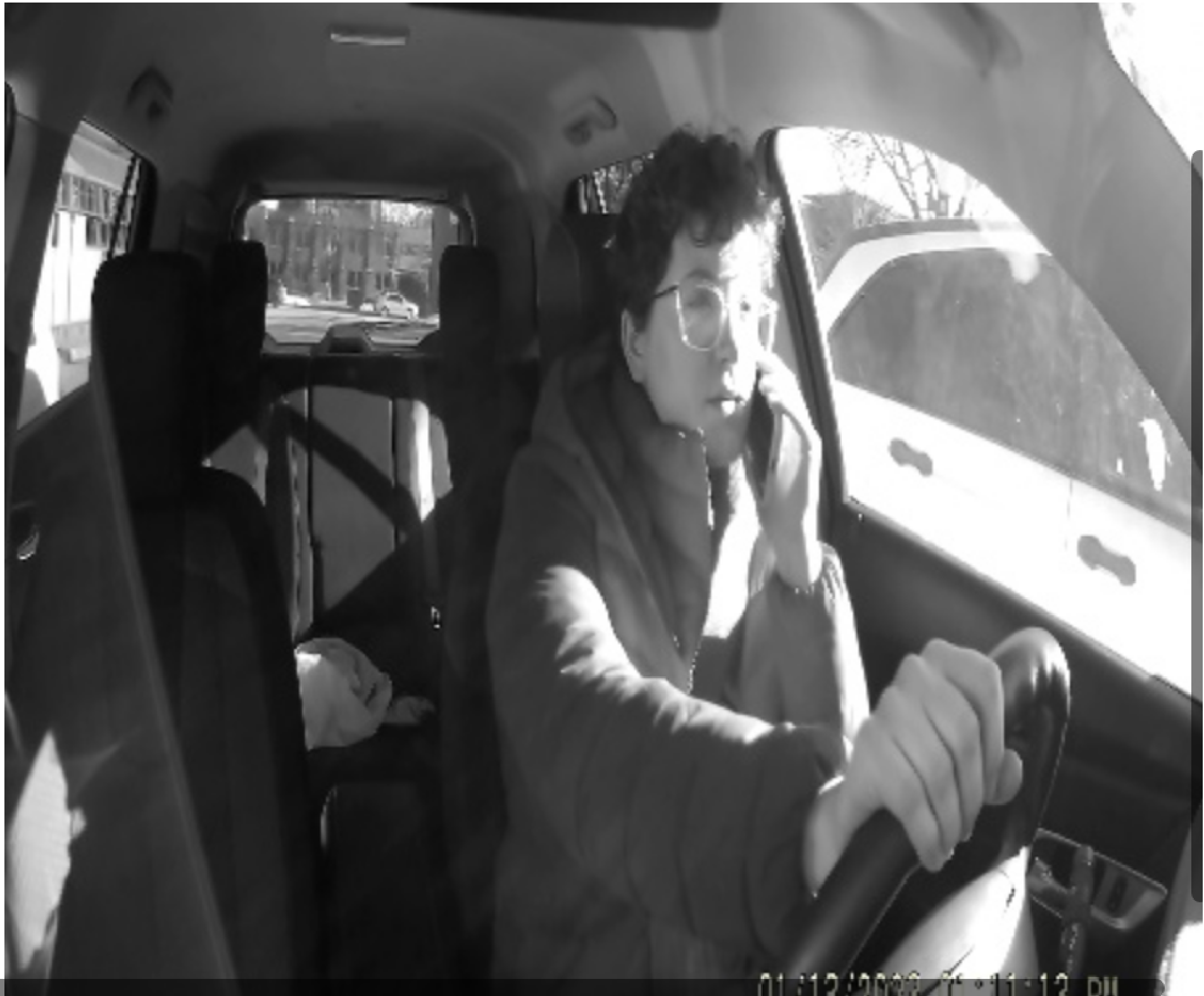}
        \caption{Rear-view}
        \label{fig:sub2}
    \end{subfigure}%
    \hfill
    \begin{subfigure}{0.15\textwidth}
        \centering
        \includegraphics[width=\linewidth,height=3cm]{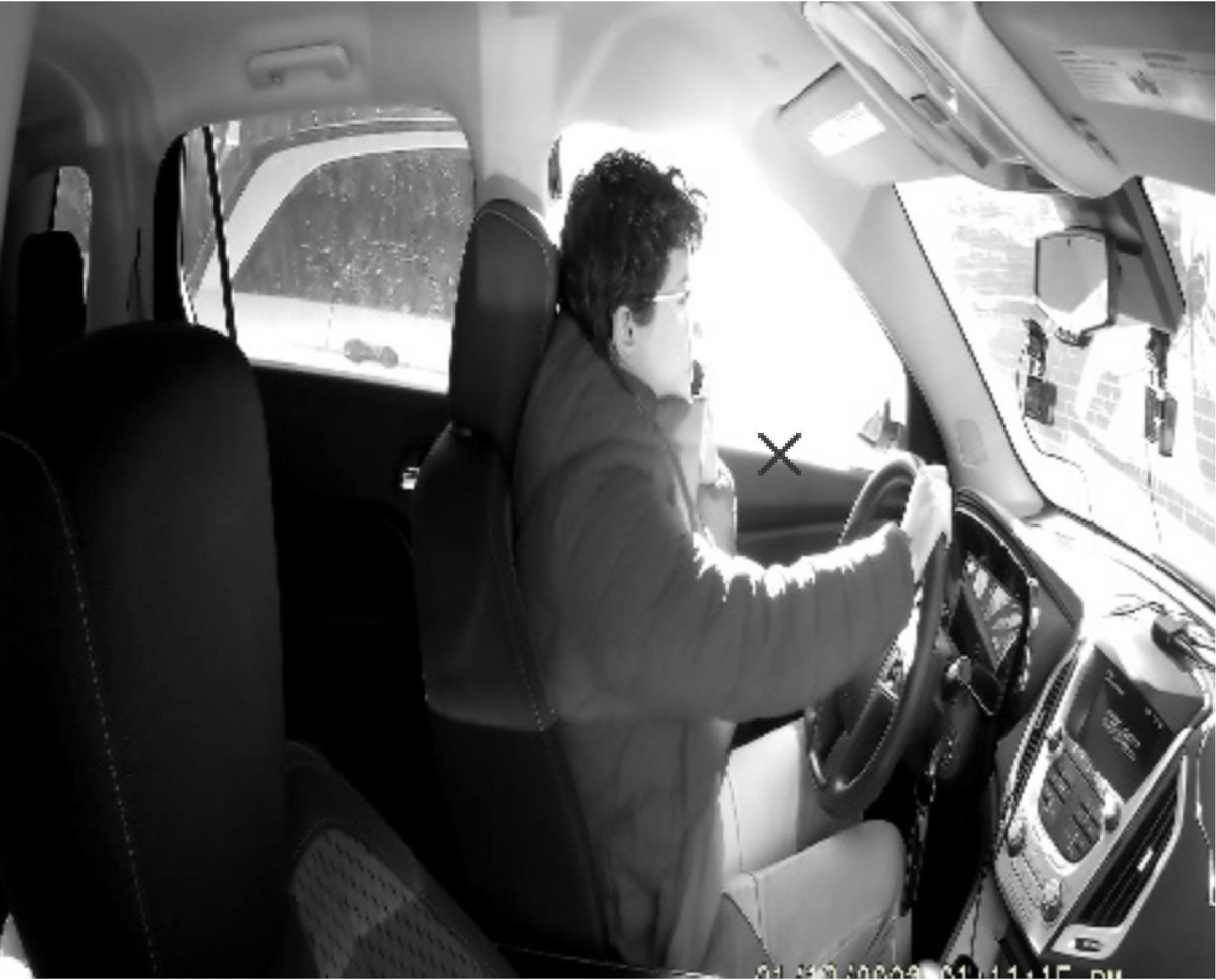}
        \caption{Right-side window}
        \label{fig:sub3}
    \end{subfigure}%
    \hfill
    \caption{Showing different camera views}
    \label{fig:camerimages}
\end{figure}

\begin{figure*}[h!]
  \centering
  \includegraphics[width=.9\linewidth]{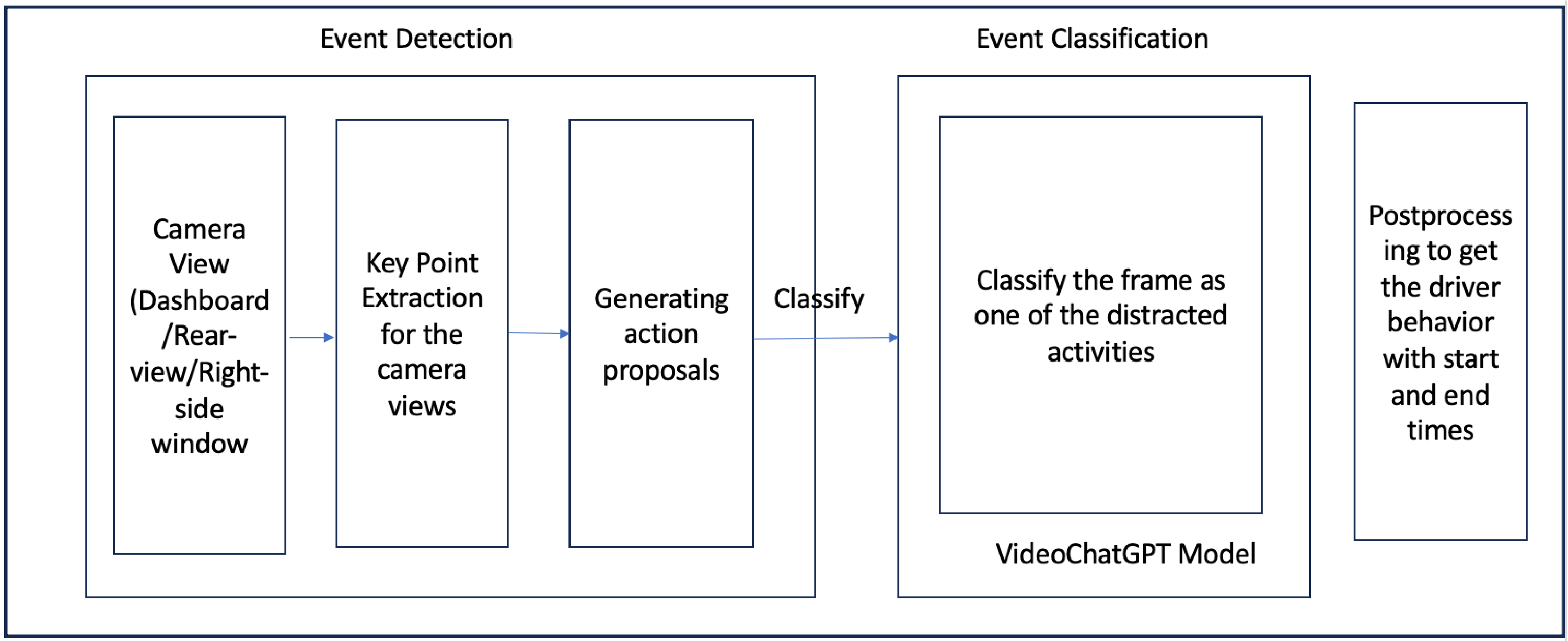}
  \caption{Showing framework pipeline}
  \label{fig:framework}
\end{figure*}

\section{Methodology}
\label{sec:methodology}

Our goal is to predict various distracted driver behaviors with their start and end times using the data from in-vehicle cameras positioned at the Dashboard, near the rearview mirror, and on the top right-side window corner (Figure \ref{fig:camerpos} shows the camera positions while Figure \ref{fig:camerimages} shows the images from the three camera views). To achieve this goal, we have subdivided it into two sub-tasks: Temporal Action Recognition (TAL), which identifies the start and end times of activities, and Activity Recognition, which recognizes the type of activity like texting, yawning, and more. Therefore, we propose an architectural framework to render these tasks.

As shown in Figure \ref{fig:framework}, our proposed framework consists of two major modules: Event detection and event classification. The Event Detection module extracts the key points related to the drivers' heads, hands, and bodies for each data frame. Using these key points, the gseg2 \cite{10.1214/14-AOS1269} algorithm finds an interval where their underlying distribution differs from the rest of the sequence. This interval serves as the start and end time of an activity performed by the driver, which is then fed as input to the event classification module to identify the type of that event. The proposed framework pipeline works as follows: feed the different intervals/events (TAL) from the first module as input to the second module for classifying those events.

We have utilized deep learning models for each module to effectively carry out the specific tasks. For the first module, we have employed YOLOv7 \cite{wang_yolov7_2023}, a pre-trained model recognized for its robust object detection capabilities and pose estimation for the key-point extraction. For the subsequent module, we have used the VideoChatGPT model \cite{maaz2023videochatgpt}.

\subsection{Event detection}
This module performs two major tasks: First, it extracts the key points from the video for each frame. Secondly, using these extracted key points, the change point detection algorithm finds an interval in the video data frames that is different from the rest of the sequence of data frames.

\subsubsection{Key-Point Extraction}
\begin{figure}[H]
  \centering
   \includegraphics[width=0.9\linewidth]{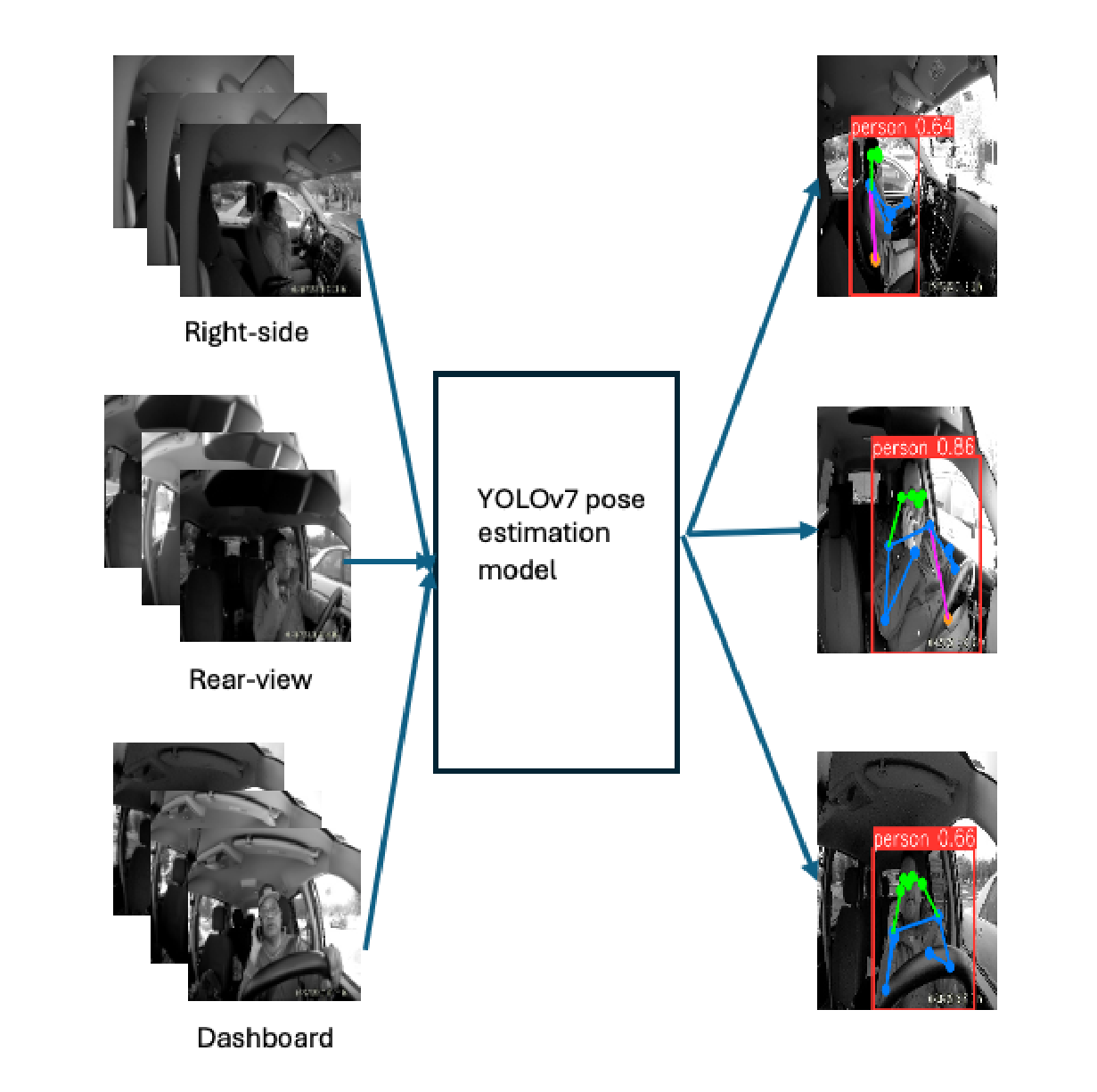}

   \caption{Showing key points extraction}
   \label{fig:keypoints}
\end{figure}

We have utilized YOLO (You Only Look Once) \cite{wang_yolov7_2023}, a renowned real-time object detection and image segmentation model, to extract the key points. Specifically, we have used yolov7 pose estimation model to identify the location of key points in each frame, along with their respective confidence scores. The locations of these points represented a set of 2D [x, y] coordinates, with confidence scores deciding their visibility. We only selected points which had confidence scores greater than 50\%. Moreover,these points were normalized as discussed in section 4.1. These key points are obtained from each frame, focusing on individuals' facial, hand, and body within the video. They play a vital role in providing essential information to the Interval Selection module, which employs these key points to find a statistically significant interval different from the given data sequence. As shown in Figure \ref{fig:keypoints}, the key points are generated for each camera view.

\subsubsection{Interval Selection}
We have executed Graph-Based Change-Point Detection for Changed Interval algorithm (gseg2) \cite{gseg2} to find intervals from the given data set. This algorithm is a part of the R package and has the following usage: 
gseg2(n, E, statistics=c("all","o","w","g","m"), l0=0.05*n, l1=0.95*n, pval.appr=TRUE, skew.corr=TRUE, pval.perm=FALSE, B=100) where 'n' is the number of observations in the sequence, 'E' is the edge matrix (a "number of edges" by two matrices) for the similarity graph. Each row contains the node indices of an edge, 'l0' is the minimum length of the interval to be considered as a changed interval, 'l1' is the maximum length of the interval to be considered as a changed interval, 'pval.appr'  if it is TRUE, the function outputs p-value approximation based on asymptotic properties, 'skew.corr' is applicable only when 'pval.appr'=TURE. If skew.corr is TRUE, the p-value approximation will incorporate skewness correction, 'pval.perm'  if it is TRUE, the function outputs p-value from doing B permutations, 'B'  is useful only when pval.perm=TRUE. The default value for B is 100.

The algorithm needed data in a specific format which is explained in section 4.2. We experimented with different values of n, l0, and l1 for our case and left others as default. We have presented the results corresponding to the best combinations. The algorithm outputs results for four statistics: Original edge-count scan statistic ("o"), Weighted edge-count statistic ("w"), Generalized edge-count statistic ("g"), and Max-type edge-count statistic ("m"). We have used Max-type edge-count statistics for finding the interval as we have key points corresponding to human posture that change with time. After getting the proposed intervals from all the camera views, we applied post-processing to select common intervals among the camera views.

\subsection{Event Classification}

 Video-LLM showcases the impressive capabilities of learning a wide range of visual concepts and exhibits exceptional performance on different few-shot tasks using a pre-trained model. We have selected VideoChatGPT, a version of Video-LLM, as our fundamental model for detecting distracted behaviors in video clips. We chose this model because of its ability to be pretty accurate with limited resources and its proven superior performance over other Video-LLMs, mainly when resources are scarce. Our strategy involves redefining event classification into a video question-answering (VQA). We trained VideoChatGPT with a comprehensive question covering all sixteen identified events to ensure the model is efficiently tailored to our needs. We also recognize the challenge posed by the small size of our dataset on distracted behaviors. It is another reason to use a pre-trained Video-LLM, which reduces the risk of overfitting during the VQA process. These strategies aim to avoid overfitting and catastrophic forgetting, maintaining the model's robustness and generalization ability. In summary, our objective is to leverage VideoChatGPT's capabilities for video understanding tasks, focusing on classifying distracted actions while effectively navigating the hurdles of limited resources and the risk of overfitting.

\section{Experiments}
\label{sec:experiments}

\subsection{SynDD2}
This dataset was introduced in the 7th AI City Challenge \cite{naphade_7th_2023}. It contains videos of distracted driving activities recorded from three in-vehicle cameras in synchronization positioned at the Dashboard, Rearview, and Right-side window. The dataset has data from participants who performed 16 distracted driving activities, as listed in Table \ref{tab:syndd2-class}.  These 16 activities have a random order,  and each of them lasts for a random duration of time. Thus, classifying each activity with its start and end times (localization) becomes challenging.

\begin{table} 
    \centering
    \footnotesize
    \begin{tabular}{|c|c|} \hline 
         ID& Activity Classes\\ \hline 
         1&  Normal Forward Driving\\ \hline 
         2&  Drinking\\ \hline 
         3&  Phone Call(right)\\ \hline 
         4& Phone Call(left)\\ \hline 
         5& Eating\\ \hline 
         6& Text (Right)\\ \hline 
         7& Text (Left)\\ \hline 
         8& Reaching behind\\ \hline 
 9&Adjust control panel\\ \hline 
 10&Pick up from floor (Driver)\\ \hline 
 11&Pick up from floor (Passenger)\\ \hline 
 12&Talk to passenger at the right\\ \hline 
 13&Talk to passenger at backseat\\ \hline 
 14&Yawning\\ \hline 
 15&Hand on head\\ \hline 
 16&Singing and dancing with music\\\hline
    \end{tabular}
    \caption{Showing SynDD2: Activity classes}
    \label{tab:syndd2-class}
    \vspace{-0.4cm}
\end{table}
Each video in the dataset has a frame rate of 30 FPS and a resolution of 1920*1080. Each video has its corresponding annotation files (for the training data set), which contain information such as activity type, start and end times of each activity type, participant's appearance, etc. 

\subsection{Experimentation: Event Detection}
Figure \ref{fig:framework} shows our overall approach. First, we extracted the key points of the videos of the Dashboard, Rearview, and Right-side camera views by using the Yolov7 pose estimation model \cite{wang_yolov7_2023}.  The model provides 17 crucial key points along with their associated confidence scores, representing the visibility of these key points related to the human posture, spanning from the head to the toes. These key points include the nose, eyes, ears, and joints (shoulders, wrists, elbows, knees, etc). For our specific case, not all key points prove valuable; hence, we opted for a subset of key points relevant to the hands and head.  
\subsubsection{Data Preprocessing}
We preprocessed the data by normalizing the key points to maintain consistency across multiple videos. This involved dividing each key point by the frame size and standardizing their scale. This was done during the key point extraction step. 
\subsubsection{Data Preparation for gseg2}
The gseg2 algorithm required an edge matrix as input, and to create that we have used KMST (K-Means Spanning Tree) algorithm. This algorithm takes the input data in matrix format and calculates the distance. We have applied Euclidean distance to compute the distance of each key point from the other.
For the given points  $\mathbf{p} = (x_1, y_1)$ and $\mathbf{q} = (x_2, y_2)$ the Euclidean distance is given by:
$$ d(\mathbf{p}, \mathbf{q}) = \sqrt{(x_1 - x_2)^2 + (y_1 - y_2)^2} $$

Since our data sets contain multiple key points, we have calculated the pairwise Euclidean distances between each pair of points in the dataset. This resulted in a distance matrix where each element represented the Euclidean distance between the corresponding pair of points in the dataset. Moreover, to find the optimal k-value for KMST, we experimented with multiple k-values and concluded that making the graph denser (higher k-value) gave better results. Table \ref{tab:kmst} shows results corresponding to various k-values. In the table, actual-start and actual-end are the ground truth times in seconds while p-start and p-end are predicted times intervals predicted by gseg2 algorithm.

\begin{table}[H]
    \centering
    \footnotesize
    \begin{tabular} {|c|c|c|c|c|} \hline 
         K value& actual-start& actual-end & p-start &p-end \\ \hline 
         10&  236&  241&  233.9& 236.067\\ \hline 
         12&  236&  241&  233.9& 236.067\\ \hline 
         15&  236&  241&  233.9& 236.067\\ \hline 
         17&  236&  241&  233.9& 236.067\\ \hline 
         19&  236&  241&  233.9& 236.067\\ \hline 
         20&  236&  241&  233.9& 236.067\\ \hline 
         23&  236&  241&  233.9& 236.067\\ \hline 
         25&  236&  241&  233.9& 236.067\\ \hline 
         26&  236&  241&  237.3& 240\\ \hline
 27& 236& 241& 237.3&240\\\hline
 28& 236& 241& 237.3&240\\\hline
 30& 236& 241& 237.3&240\\\hline
    \end{tabular}
    \caption{Showing k-values with predicted intervals}
    \label{tab:kmst}
\end{table}
Finally, this matrix is fed to the gseg2 algorithm, which outputs the statistically significant change point interval. Subsequently, the interval serves as a distracted activity's start and end time, which is classified by the next module.

\subsection{Experimentation: Event Classification}
For VQA, using prompt engineering, we have found the following question works the best for our task. In this process, we have crafted three questions, and they are :

1. "Based on the following activities: Normal Forward Driving, Drinking, Phone Call (right), Phone Call (left), Eating, Text (Right), Text (Left), Reaching behind, Adjusting control panel, Pick up from floor (Driver), Pick up from floor (Passenger), Talk to passenger at the right, Talk to passenger at backseat, Yawning, Hand on head, and Singing or dancing with music, which activity is being performed in the video?"

2."Is the driver simulating any of the following activities? 1. Normal Forward Driving, 2. Pretending to drink a beverage, 3. Simulating a phone call with the right hand, 4. Simulating a phone call with the left hand, 5. Pretending to eat food, 6. Simulating texting with the right hand, 7. Simulating texting with the left hand, 8. Pretending to reach behind the seat, 9. Simulating adjusting the control panel, 10. Pretending to pick up an object from the floor on the driver's side, 11. Pretending to pick up an object from the floor on the passenger's side, 12. Simulating talking to a passenger seated on the right side, 13. Simulating talking to a passenger seated in the backseat, 14. Simulating yawning, 15. Pretending to place a hand on the head, 16. Simulating singing or dancing to music. Please provide a 'yes' or 'no' response for each activity."

3."Is the driver simulating any of the following activities? 1. Normal Forward Driving, 2. Pretending to drink a beverage, 3. Simulating a phone call with the right hand, 4. Simulating a phone call with the left hand, 5. Pretending to eat food, 6. Simulating texting with the right hand, 7. Simulating texting with the left hand, 8. Pretending to reach behind the seat, 9. Simulating adjusting the control panel, 10. Pretending to pick up an object from the floor on the driver's side, 11. Pretending to pick up an object from the floor on the passenger's side, 12. Simulating talking to a passenger seated on the right side, 13. Simulating talking to a passenger seated in the backseat, 14. Simulating yawning, 15. Pretending to place a hand on the head, 16. Simulating singing or dancing to music. Please provide the activity the driver is doing."

We utilized this question for all the videos we trimmed based on the start and end time calculated from our Key Point Change Detection process. The testing dataset has 30 videos comprising 450 activities, comprising 15 activities per video. The model answered every activity based on the start and end time. If the model detects no activity from the list provided in the question, it responds with "no." We verified the accuracy of each question using the ground truth.
\subsection{Evaluation metric.}
We have evaluated our approach using the AICITY CHALLENGE evaluation \cite{AICITYeval}.  The AICITY evaluation is defined as follows: Given a ground-truth activity \textit{g} with start time \textit{gs} and end time \textit{ge}, the aim is to find its closest predicted activity match as that predicted activity \textit{p} belonging to the same class as \textit{g} with the highest overlap score \textit{os}. Additional condition is imposed that start time \textit{ps} and end time \textit{pe} fall within the range [\textit{gs} – 10s, \textit{gs} + 10s] and [ge – 10s, ge + 10s], respectively. The overlap between \textit{g} and \textit{p} is calculated as the ratio between the time intersection and the time union of the two activities, i.e.,
\begin{figure}[H]
  \centering
   \includegraphics[width=0.8\linewidth]{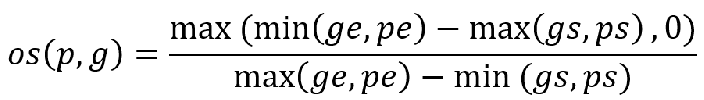}

   \caption{showing overlap score}
   \label{fig:onecol}
\end{figure}
\section{Results}
\label{sec:results}
\subsection{Performance of  event detection}
This module's main objective is to predict an activity's intervals, start and end times. We experimented with the gseg2 algorithm for different values of n (number of data samples), l0, and l1, as shown in Table \ref{tab:gseg_results}. The results are displayed for 30 video samples, which had a total of 450 activity intervals. 
\begin{table}[H]
    \centering
    \footnotesize
    \begin{tabular} 
    {|c|c|c|c|} 
    \hline 
         n-data samples&Accurate predictions&  Accuracy \%& L0, L1\\ \hline 
         1 minute &  242&  51& 0.1, 0.90 \\ \hline 
         2 minutes&  235&  49& 0.1, 0.90 \\ \hline 
         3 minutes&  228&  47& 0.1, 0.90 \\ \hline 
         4 minutes&  176&  42& 0.1, 0.90 \\ \hline
    \end{tabular}
    
    \caption{Showing results for different n}
    \label{tab:gseg_results}
\end{table}
We have used the data samples from 1, 2, 3, and 4 minutes, and to address scenarios where activities extend beyond these time intervals, such as an activity commencing just before the 1-minute and continuing into the next minute, we re-executed the algorithm using data samples starting from 30 seconds for 1 minute. Similar approaches were applied for other duration's to accommodate boundary conditions.

We have evaluated our approach using the AICITY CHALLENGE evaluation \cite{AICITYeval}. Although the AICITY CHALLENGE evaluation is designed for final results (activity recognition with start and end times), we have leveraged it to assess the accuracy of intervals. This unconventional use of the evaluation metric aims to ensure our approach's robustness. We reasoned that if our approach accurately predicted start and end times, it would likely also enable the activity recognition algorithm to identify activity types effectively.

\begin{figure*}[ht!]
    \centering
    \begin{subfigure}{0.24\textwidth}
        \centering
        \includegraphics[width=\linewidth,height=3cm]{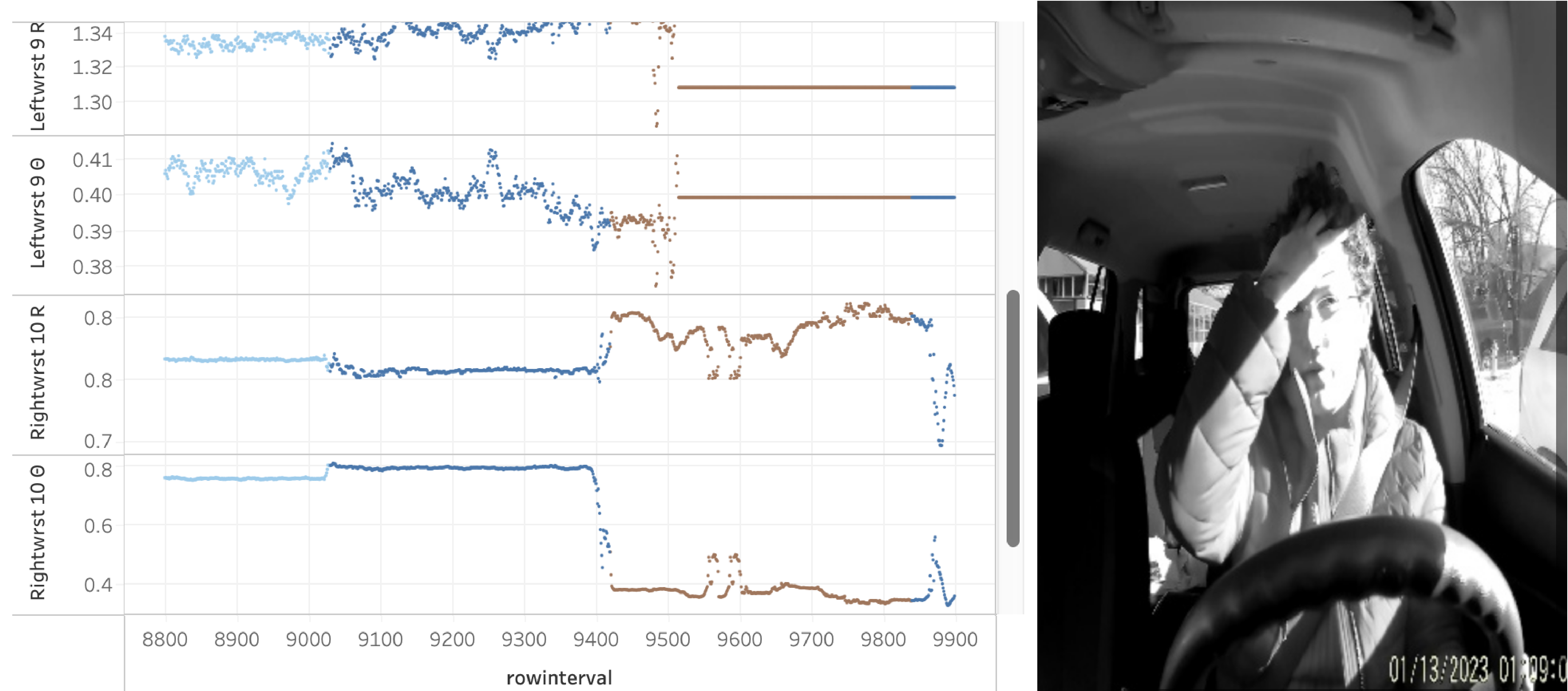}
        \caption{Hand on Head}
        \label{fig:sub1}
    \end{subfigure}%
    \hfill
    \begin{subfigure}{0.24\textwidth}
        \centering
        \includegraphics[width=\linewidth,height=3cm]{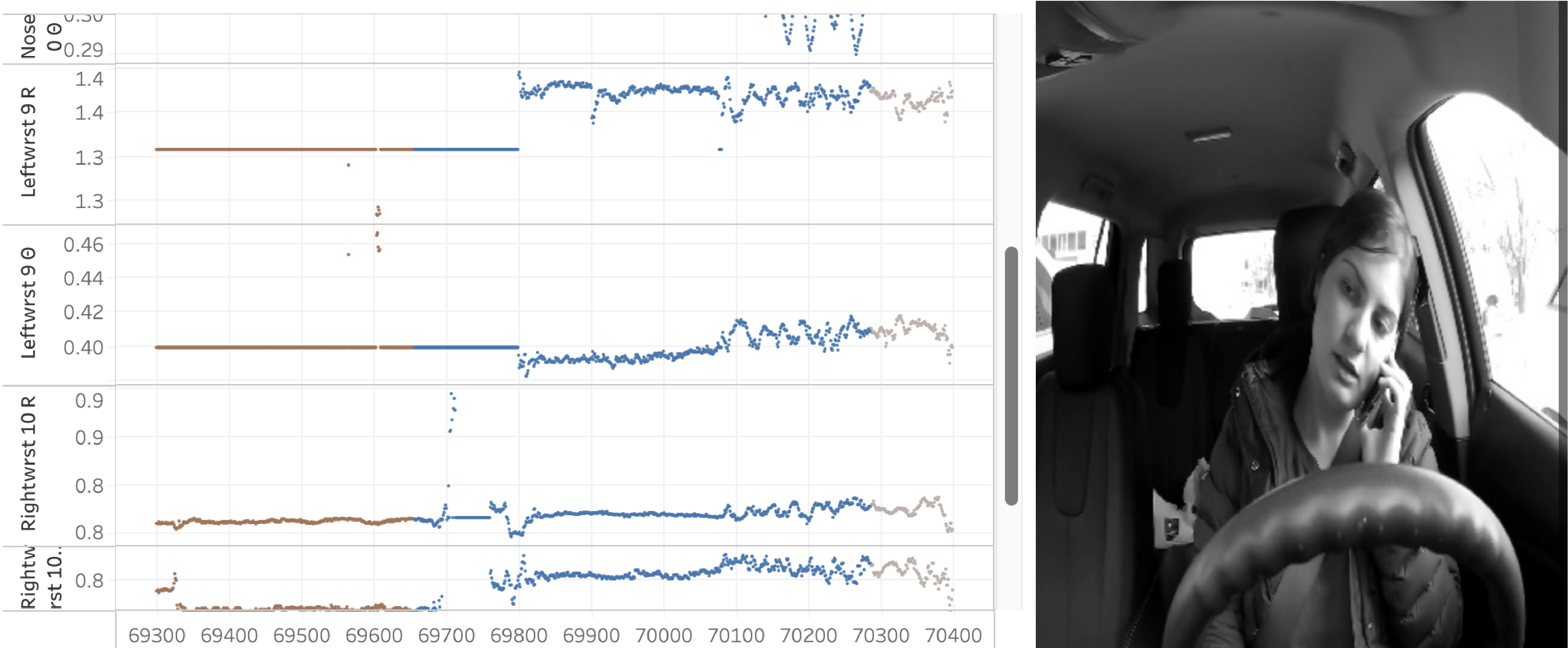}
        \caption{Phone call-left}
        \label{fig:sub2}
    \end{subfigure}%
    \hfill
    \begin{subfigure}{0.24\textwidth}
        \centering
        \includegraphics[width=\linewidth,height=3cm]{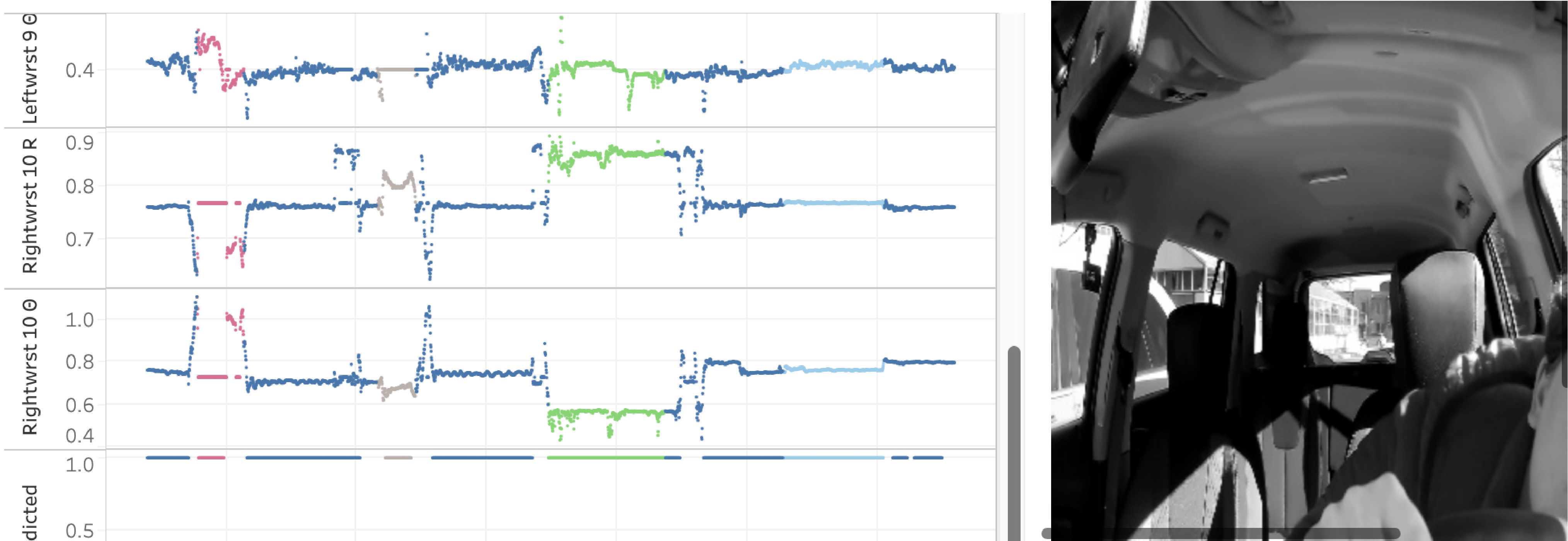}
        \caption{Picking from floor}
        \label{fig:sub3}
    \end{subfigure}%
    \hfill
    \begin{subfigure}{0.24\textwidth}
        \centering
        \includegraphics[width=\linewidth,height=3cm]{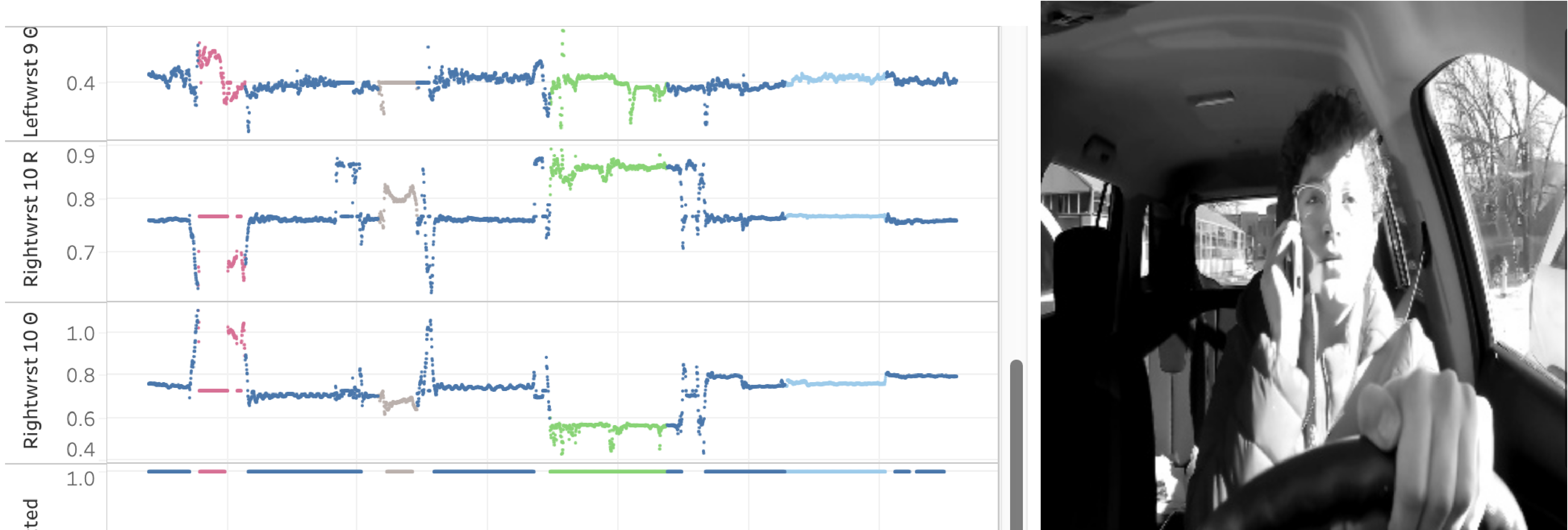}
        \caption{Phone call-right}
        \label{fig:sub4}
    \end{subfigure}
    \caption{Showing few classes where the change-point detection produced correct action proposals}
    \label{fig:combined_pos}
\end{figure*}

\begin{figure*}[ht!]
    \centering
    \begin{subfigure}{0.24\textwidth}
        \centering
        \includegraphics[width=\linewidth,height=3cm]{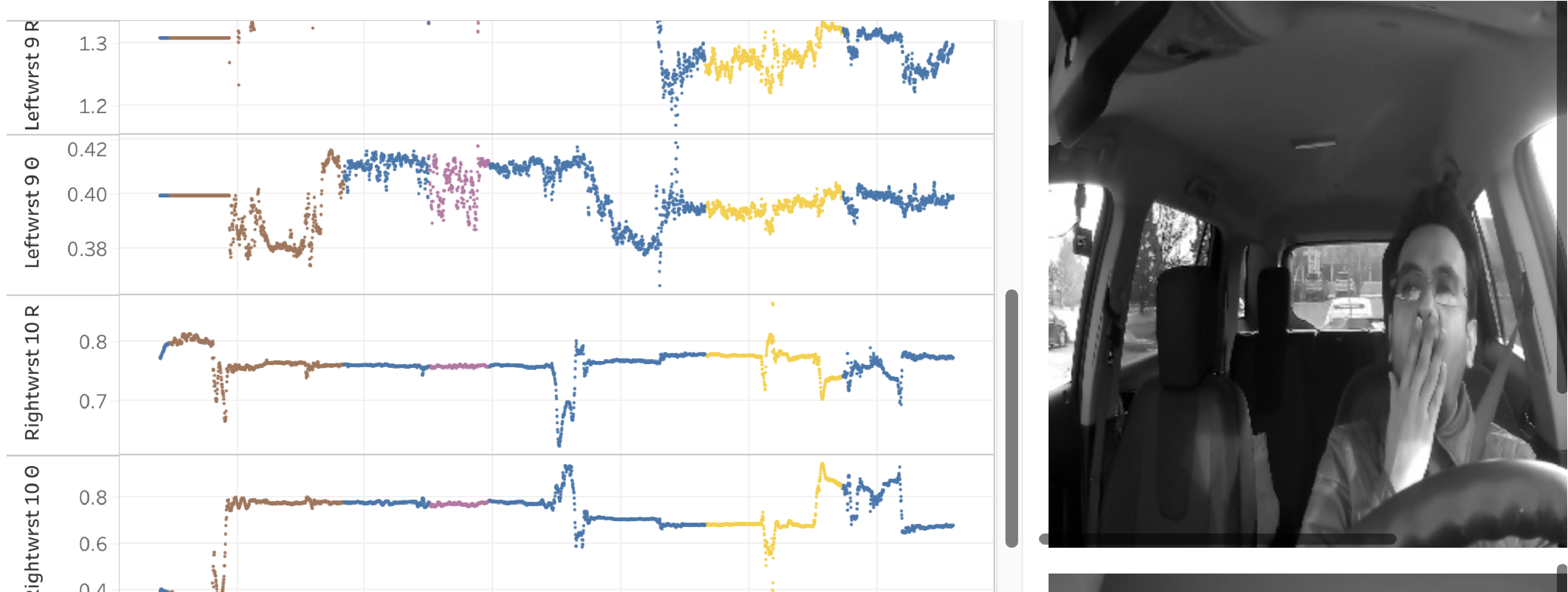}
        \caption{Yawning}
        \label{fig:sub1}
    \end{subfigure}%
    \hfill
    \begin{subfigure}{0.24\textwidth}
        \centering
        \includegraphics[width=\linewidth,height=3cm]{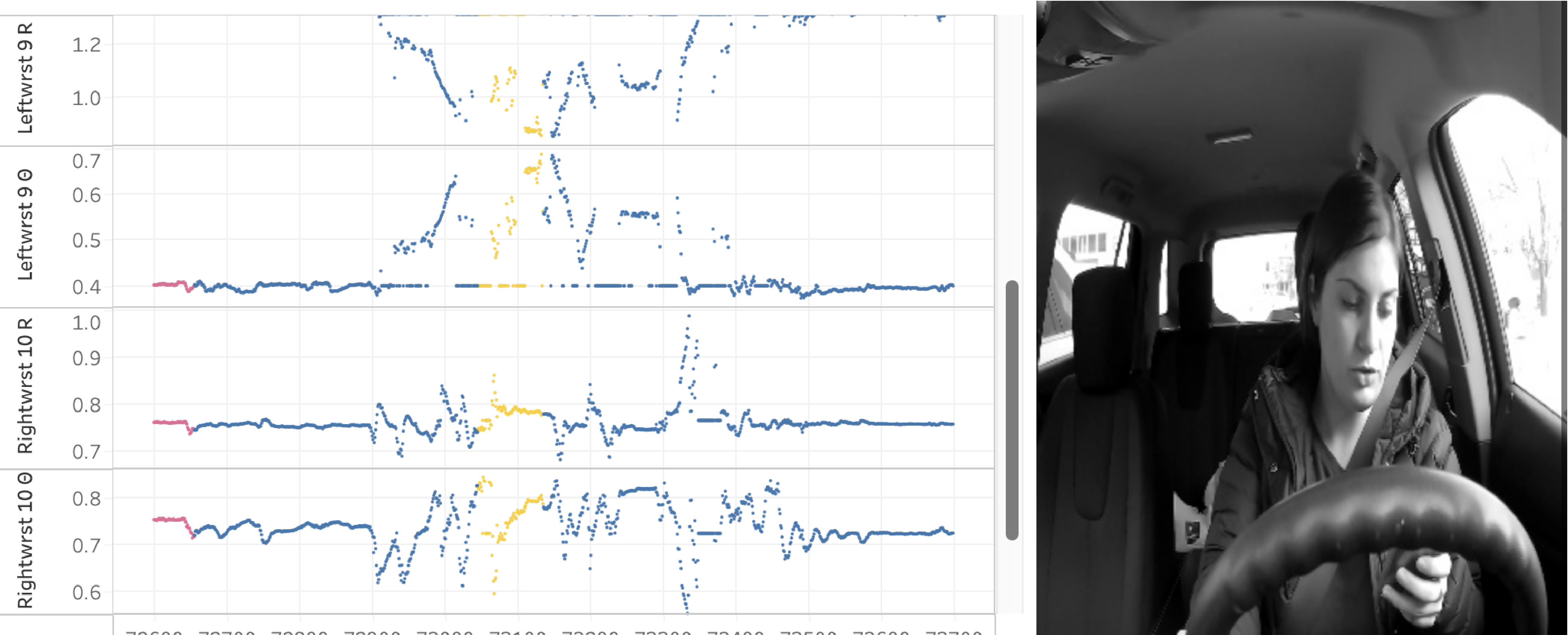}
        \caption{Text left}
        \label{fig:sub2}
    \end{subfigure}%
    \hfill
    \begin{subfigure}{0.24\textwidth}
        \centering
        \includegraphics[width=\linewidth,height=3cm]{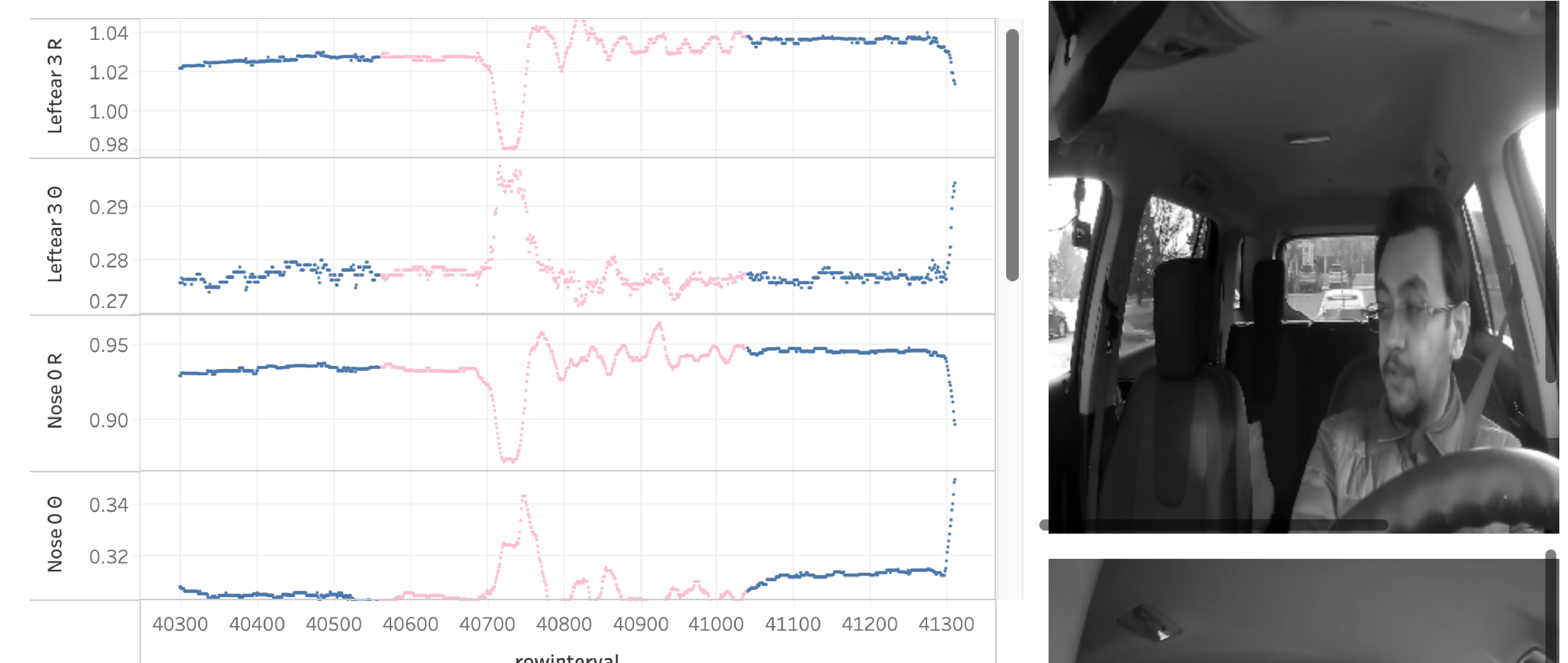}
        \caption{Talking}
        \label{fig:sub3}
    \end{subfigure}%
    \hfill
    \begin{subfigure}{0.24\textwidth}
        \centering
        \includegraphics[width=\linewidth,height=3cm]{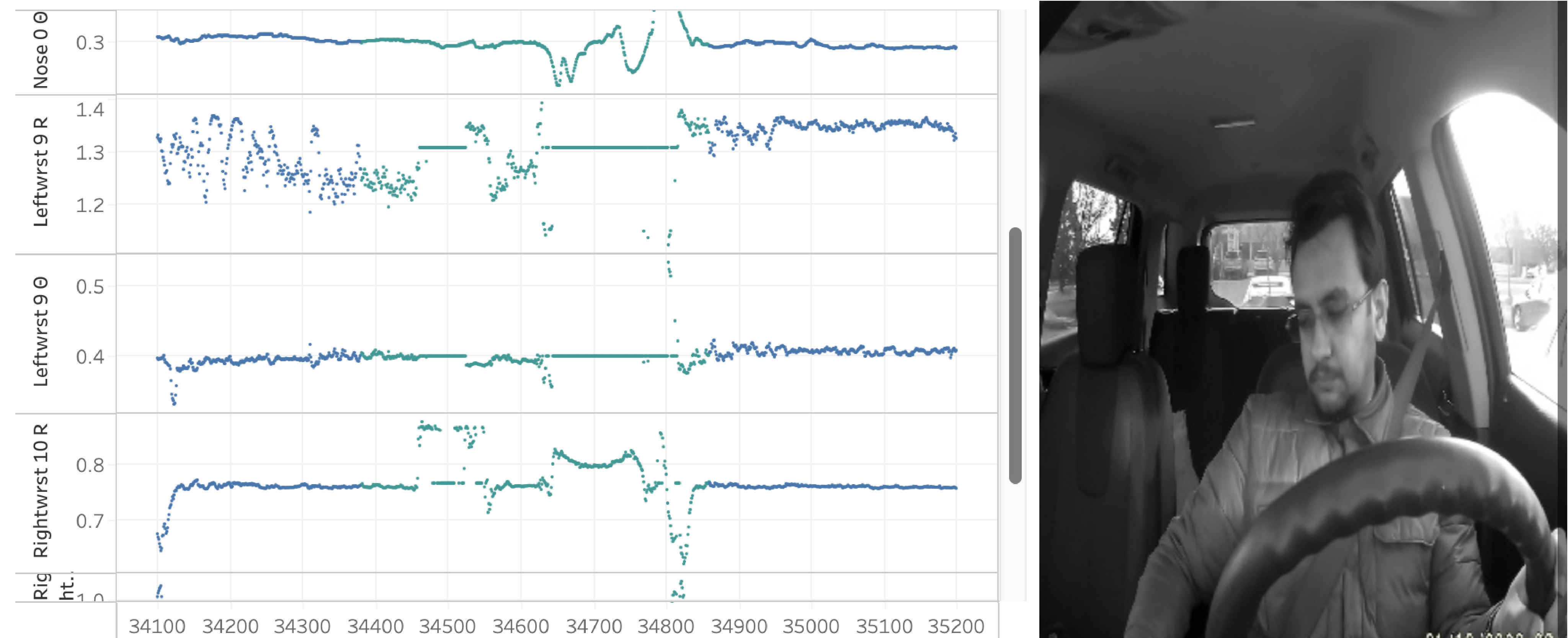}
        \caption{Text right}
        \label{fig:sub4}
    \end{subfigure}
    \caption{Showing few classes where the change-point detection could not produced correct action proposals}
    \label{fig:combined_neg}
\end{figure*}

We have specifically used the condition which states that start time \textit{ps} and end time \textit{pe} are in the range [\textit{gs} – 10s, \textit{gs} + 10s] and [ge – 10s, ge + 10s], respectively, to find the accuracy of the predicted intervals. We used this approach to find the accuracy of action proposals of the last years top performing team, the results are shown in Table \ref{tab:proposal_accuracy}.

\begin{table}[H]
    \centering
    \footnotesize
    \begin{tabular}{|c|c|} \hline 
         Approach& Accuracy \%\\ \hline 
         Meituan-IoTCV \cite{zhou2023multi} & 40\\ \hline 
         change-point-detection (\textbf{ours})& 51\\ \hline
    \end{tabular}
    \caption{Showing action proposal prediction accuracy}
\label{tab:proposal_accuracy}
\end{table}

\subsubsection{Error Analysis of the change point algorithm}
The algorithm have worked better for few classes while it  failed for others. Specifically, the algorithm picks the classes that involves movements of hands or body by a large amount. For example the classes "Hand on head" ,"phone calls", "picking from floor" is mostly predicted by the algorithm while "texting" "yawning", "singing","talking" is not predicted most of the time.  One of the reason for this is the presence of noise in the data (key points) which makes it difficult to differentiate anomaly from normal. 

Figure \ref{fig:combined_pos} shows the classes where the algorithm performed well. The changes in key points are visible for the left hand and the right hand. Similarly, Figure \ref{fig:combined_neg} shows classes where the model didn't perform well because of the noise in the data or because the activity didn't involve much movement of hand or head. For example, "talking", "singing" and "yawning" do not involve any movements. 

Overall, the change point algorithm worked well because it was able to locate the activity start and times given that there was noise in the data and some of the activities did not cause any changes in the key points.

\subsection{Performance of  event classification}
We conducted experiments to evaluate the performance of our event classification module using VideoChatGPT, which is a tailored version of Video-LLM. VideoChatGPT is designed to detect distracted behaviors in video clips and followed the exact same way it is in the official repository. To assess the accuracy of the model for classifying the identified distracted driving activities, we crafted three comprehensive questions using prompt engineering in previous section.  The questions cover all sixteen activities and are designed to evaluate the model's ability to classify the activities accurately based on start and end times determined by the Key Point Change Detection process. AS we mentioned before, the testing dataset consisting of 30 videos, each containing 15 activities, resulting in a total of 450 activities. For each activity, the model provided a response based on the specified time intervals. If the model was unable to detect any activity from the provided list, it responded with "no." We verified the accuracy of each question against the ground truth.

\begin{table}[H]
\centering
\footnotesize
\begin{tabular}{|c|c|c|}
\hline
Question Number&Accurate Predictions&Accuracy(\%) \\ \hline
1 & 223 & 49.5 \\ \hline
2 &249  &55.3  \\ \hline
3 &259  &57.5 \\ \hline
\end{tabular}
\caption{Showing Model Performance on three questions}
\label{tab:model_performance}
\end{table}

In Table \ref{tab:model_performance}, it is clear that the accuracy peak was reached with the third question, achieving an impressive 57.5\%. This milestone underscores the substantial impact of our refined prompt engineering technique on bolstering VideoChatGPT's classification capabilities. The discernible upward trend in accuracy rates from the initial to the third question highlights the model's enhanced comprehension of the tasks at hand, thanks to the iterative refinement of the prompts. These findings underscore VideoChatGPT's promising utility in accurately classifying distracted driving behaviors, showcasing its potential in real-world applications. To the best of our knowledge, this is the first time Video-LLMs have been used for this task. Therefore, we did not report a baseline score.

\subsection{Conclusion}
Our DeepLocalization model is an effective solution for detecting and locating distracted driver behavior in real time. The approach we use involves key-point extraction, change-point detection, and video language modeling to precisely identify and temporally localize a wide range of driver activities. Our experimental results on the SynDD2 dataset demonstrate the effectiveness of our approach in addressing the complex problem of distracted driving with limited resources. In the future, to improve the accuracy of event detection and classification, we plan to introduce a more robust key point detection algorithm, which is currently the bottleneck in achieving better accuracy. Additionally, fine-tuning for one or two epochs is a viable way to improve accuracy even on a consumer-grade GPU.

{
    \small
    \bibliographystyle{ieeenat_fullname}
    \bibliography{main}
}


\end{document}